\documentclass[letterpaper]{article} 
\usepackage{aaai2026}  
\usepackage{times}  
\usepackage{helvet}  
\usepackage{courier}  
\usepackage[hyphens]{url}  
\usepackage{graphicx} 
\usepackage{natbib}  
\usepackage{caption} 
\usepackage{algorithm}
\usepackage{algorithmic}

\usepackage{amsmath}
\usepackage{booktabs}
\usepackage{cleveref}
\usepackage{adjustbox}
\usepackage{multirow}
\usepackage{pifont}
\usepackage{color, colortbl}
\usepackage{amsfonts} 
\usepackage{comment}
\usepackage{subcaption}
\usepackage{array}
\usepackage{tikz} 
\newcolumntype{C}[1]{>{\centering\arraybackslash}p{#1}}
\usepackage{url}

\usepackage{microtype}
\definecolor{delta_color}{HTML}{16C47F}

\usepackage{newfloat}
\usepackage{listings}
\DeclareCaptionStyle{ruled}{labelfont=normalfont,labelsep=colon,strut=off} 
\floatstyle{ruled}
\newfloat{listing}{tb}{lst}{}
\floatname{listing}{Listing}
\title{PaSE: Prototype-aligned Calibration and Shapley-based Equilibrium for Multimodal Sentiment Analysis}
\author{
    Kang He\textsuperscript{\rm 1,2},
    Boyu Chen\textsuperscript{\rm 1},
    Yuzhe Ding\textsuperscript{\rm 1},
    Fei Li\textsuperscript{\rm 1},
    Chong Teng\textsuperscript{\rm 1},
    Donghong Ji\textsuperscript{\rm 1}\thanks{Corresponding author}
}
\affiliations{
    \textsuperscript{\rm 1}
    Key Laboratory of Aerospace Information Security and Trusted Computing, Ministry of Education, \\School of Cyber Science and Engineering, Wuhan University \\
    \textsuperscript{\rm 2} 
    Shanghai Innovation Institute

    {\{hekang0225, chenboyu1225, yuzheding, lifei\_csnlp, chongteng, dhji\}@whu.edu.cn}\\
}

\usepackage{bibentry}

\begin{document}

{\maketitle
\begin{abstract}

Multimodal Sentiment Analysis (MSA) seeks to understand human emotions by integrating textual, acoustic, and visual signals.
Although multimodal fusion is designed to leverage cross-modal complementarity, real-world scenarios often exhibit \textit{\textbf{modality competition}}: dominant modalities tend to overshadow weaker ones, leading to \textit{suboptimal performance}.
In this paper, we propose PaSE, a novel \textbf{P}rototype-\textbf{a}ligned Calibration and \textbf{S}hapley-optimized \textbf{E}quilibrium framework, which enhances collaboration while explicitly mitigating modality competition.
PaSE first applies Prototype-guided Calibration Learning (PCL) to refine unimodal representations and align them through an Entropic Optimal Transport mechanism that ensures semantic consistency.
To further stabilize optimization, we introduce a Dual-Phase Optimization strategy.
A prototype-gated fusion module is first used to extract shared representations, followed by Shapley-based Gradient Modulation (SGM), which adaptively adjusts gradients according to the contribution of each modality.
Extensive experiments on IEMOCAP, MOSI, and MOSEI confirm that PaSE achieves the superior performance and effectively alleviates modality competition. 

\maketitle
\end{abstract}

\section{Introduction}

Multimodal Sentiment Analysis (MSA) aims to infer human emotions by integrating information from diverse modalities such as text, audio, and visual data. Compared to traditional unimodal approaches \cite{fan-etal-2022-sentiment,wang-etal-2024-refining}, MSA leverages the complementary nature of modalities to achieve deeper emotional understanding \cite{poria2016convolutional,zhu2024kebr,xu2024leveraging}. 
With the rapid progress in representation learning \cite{10887581,he-etal-2025-dalr} and the proliferation of multi-source data, MSA has become a research hotspot, finding applications in dialogue systems \cite{zhang-etal-2024-escot}, stance detection \cite{ding-etal-2025-zero} and social media analysis \cite{zhai-etal-2024-chinese}.

A central challenge in MSA lies in effectively fusing heterogeneous signals to optimize cross-modal interactions. Early works focused on feature-level concatenation \cite{poria2016convolutional,yu2019adapting}, while subsequent efforts explored decision-level fusion by aggregating predictions from individual modalities \cite{zhu2024kebr,zhuang2024glomo}. These paradigms generally presume that \textit{modality collaboration dominates the fusion process} and that combining modalities will inherently yield performance gains.

However, real-world MSA tasks often involve a \textit{dynamic trade-off between modalities}. Variations in data distribution, representational capacity, convergence rate, and noise levels can result in imbalanced learning across modalities \cite{wang2020makes,fujimori2020modality,huang2022modality}. Such imbalance may lead to suboptimal fusion outcomes, where the benefits of multimodal integration do not meet expectations \cite{zhou2023intra}. As a result, the ideal of \textbf{cross-modal Pareto optimality} is rarely realized in practice.

\begin{figure}[t]
    \centering \includegraphics[width=1.00\columnwidth]{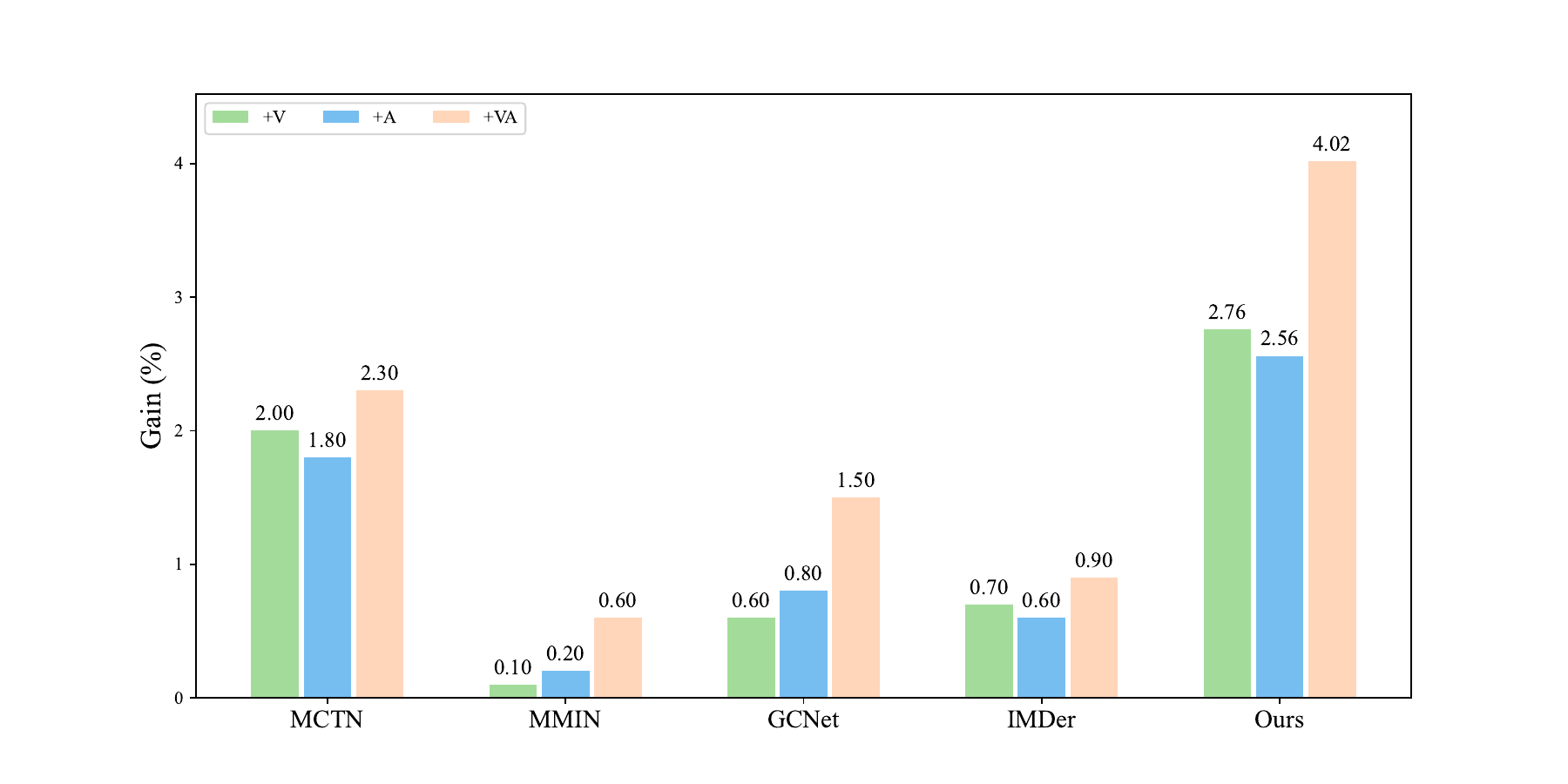}
    \caption{
   Performance improvements (F1 score) from adding audio (+A), visual (+V), or both modalities (+VA) to text-only baselines on the CMU-MOSI datasets.}
    \label{fig:intro_compare}
\end{figure}

\citet{huang2022modality} observed that unimodal models can occasionally outperform their multimodal counterparts—an effect termed \textbf{\textit{modality competition}}. As shown in Figure~\ref{fig:intro_compare}, incorporating audio and visual modalities into a text-only baseline yields limited performance gains, suggesting that modality interactions may involve interference or rivalry rather than straightforward cooperation. Recent studies \cite{wu2022characterizing,fan2023pmr,zhou2023intra} further characterize multimodal learning as a non-cooperative game, wherein dominant modalities exert greater influence by transmitting stronger gradient signals \cite{peng2022balanced}, suppressing the contributions of weaker modalities.
While this phenomenon has garnered growing attention, the intricate dynamics between modality collaboration and competition remain underexplored, posing a persistent challenge in MSA. These issues underscore the need for more adaptive fusion strategies that explicitly address modality imbalance and competition while promoting complementary integration.

To address the challenges of modality competition, we propose PaSE, a prototype-guided fusion framework that promotes cross-modal synergy while mitigating dominance effects. PaSE first calibrates unimodal features via Prototype-guided Calibration Learning (PCL), then aligns cross-modal semantics through Entropic Optimal Transport-based alignment learning (CAL). 
To further ensure balanced learning, we introduce a Dual-Phase Optimization strategy. In the first phase, Prototype-Gated Fusion mechanism (PGF) integrates modalities to learn a robust overall representation, allowing dominant modalities to guide early training. 
In the second phase, Shapley-based Gradient Modulation (SGM) dynamically reweights gradients based on each modality’s marginal contribution, mitigating long-term imbalance. 
Our main contributions are summarized as follows:

\begin{enumerate}
    \item We propose PaSE, a multimodal framework that integrates prototype-aligned calibration with Shapley-optimized equilibrium to address modality competition in MSA.
    
    \item We introduce an entropic optimal transport-based cross-modal prototype alignment mechanism that enforces structural consistency and bidirectional matching, effectively mitigating semantic discrepancies across modalities.
    
    \item We develop a dynamic fusion mechanism with Shapley-based gradient modulation to adaptively balance modality contributions and improve collaborative optimization.

    \item Extensive experiments on three benchmark datasets (IEMOCAP, MOSI, and MOSEI) demonstrate that PaSE achieves state-of-the-art performance and effectively mitigates modality competition. 

\end{enumerate}

\section{Related Work}

\subsection{Multimodal Sentiment Analysis}

Fusion is a central challenge in MSA \cite{yang2020cm, mai2022hybrid}. Early studies concatenate features from multiple modalities for joint representation learning \cite{poria2016convolutional,yu2019adapting}, while later approaches adopt decision-level fusion by aggregating unimodal predictions \cite{zhu2024kebr,zhuang2024glomo}.
With the rise of attention mechanisms, cross-modal attention has become popular for adaptive fusion \cite{tsai2019multimodal,yang2020cm,mai2022hybrid}. Another line of work explores modality disentanglement. MISA \cite{hazarika2020misa} separates shared and private representations via dedicated encoders. Follow-up methods incorporate this idea with distillation \cite{li2023decoupled} and contrastive learning \cite{yang2023confede} to boost cross-modal interaction.
Despite these advances, most methods neglect the adverse effects of modality competition. To address this, we propose a prototype-guided calibration alignment and a Shapley-based gradient modulation mechanism to balance inter-modal collaboration and competition.

\subsection{Modality Competition}

Modality imbalance \cite{fernando2021missing,peng2022balanced,fan2023pmr} arises from differences in information quality and data distribution, often causing dominant modalities to suppress weaker ones during training. Prior solutions include gradient modulation \cite{peng2022balanced,fernando2021missing}, adaptive learning rates \cite{wang2020makes,wu2022characterizing}, and modality-specific objectives \cite{fan2023pmr,fan2024detached}.
However, most rely on indirect signals such as gradient norms or losses, lacking principled quantification of each modality’s contribution. We address this by introducing Shapley values to estimate marginal utility and guide gradient modulation, enabling more effective mitigation of modality competition.

\begin{figure*}[t]
    \centering \includegraphics[width=2.0\columnwidth]{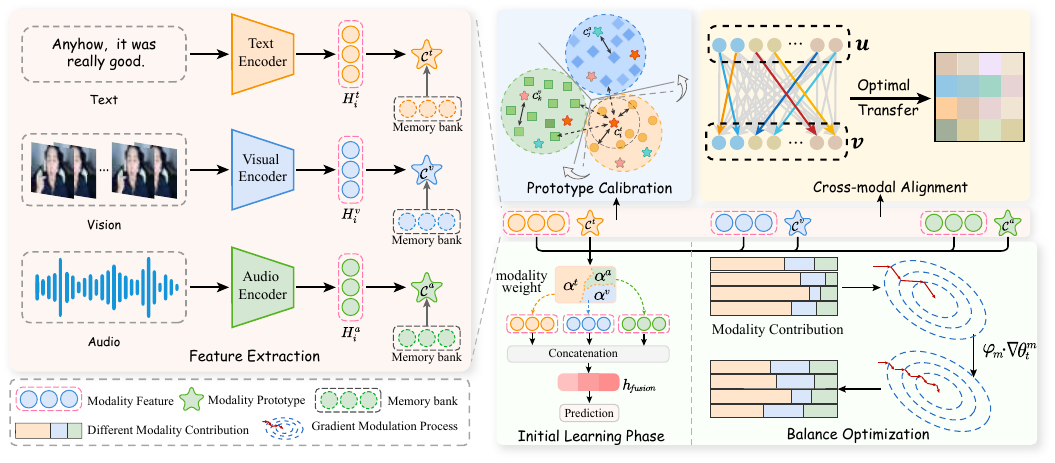}
    \caption{
    The overall architecture of our proposed model PaSE.}
    
    \label{fig:model}
\end{figure*}

\section{Task Setup}
MSA aims to predict the emotional polarity (e.g., positive, neutral, negative) or emotional intensity of a given utterance by integrating information from multiple modalities, such as text ($t$), audio ($a$), and vision ($v$). Formally, for each modality, we represent the sequence of each modality as $H^{m}\in\mathcal{R}^{B_{m}\times d_{m}}$, where $m \in \mathcal{M}=\{t,a,v\}$ represents different modalities, $B_m$ and $d_m$ denote the corresponding dimensions and sequence lengths.

\section{Methodology}
As illustrated in Figure~\ref{fig:model}, PaSE consists of three main modules: namely the intra-modal prototype calibration learning (PCL), the cross-modal alignment learning (CAL), and the dual-phase optimization module. 
The latter is further divided into the modality fusion module and the Shapley-based gradient modulation (SGM). Detailed descriptions are provided in the following subsections.

\subsection{Intra-modality Prototype Calibration (PCL)}
As previously noted, imbalanced expressive power across modalities often causes weaker modalities to be overshadowed by dominant ones \cite{wu2022characterizing,fan2023pmr}, hindering effective fusion and degrading overall performance \cite{zhou2023intra}.
To mitigate this, we introduce a Prototype-guided Calibration Learning (PCL) strategy. The key idea is to leverage category prototypes to guide weaker modalities toward a more discriminative and structured feature space, thereby avoiding suppression by stronger modalities. This also facilitates better modeling of both intra-class cohesion and inter-class separability within each modality.
Each category prototype is defined as the centroid of features belonging to that category in the corresponding modality:

{\small
\begin{equation}
    c_k^m=\frac{1}{|B_k|}\sum_{i\in B_k}h_i^m
\end{equation}
}where $c_k^m$ is the prototype of category $k$ in modality $m$, $B_k$ is the set of samples belonging to category $k$ in the mini-batch. $h_i^m$ is the feature representation of sample $i$ in modality $m$. 
To ensure stability during prototype updates, we adopt a momentum-based moving average strategy \cite{he2020momentum}:
{\small
\begin{equation}
    c_{k}^{m}\leftarrow\gamma c_{k}^{m}+(1-\gamma)\frac{1}{|B_{k}|}\sum_{i\in\mathcal{B}_{k}}h_{i}^{m}
\end{equation}
}

To encourage samples within the same category to form a compact and consistent distribution in each modality, we introduce the intra-modal prototype calibration loss $\mathcal{L}_{intra}^{m}$:
{\small
\begin{equation}
\label{eq:3}
    \mathcal{L}_\mathrm{intra}^{m}=-\frac{1}{N}\sum_{i=1}^{N}\log\frac{e^{(\phi(h_{i}^{m},c_{y_{i}}^{m})/\tau)}}{\sum_{k=1}^{K}e^{(\phi(h_{i}^{m},c_{k}^{m})/\tau)}}
\end{equation}
}where $c_{y_{i}}^{m}$ is the prototype vector for the category of sample $i$ in modality $m$, $\phi(h_{i}^{m},c_{k}^{m})$ is the similarity function, $\tau$ is the temperature parameter, and $N$ is the batch size. 
This loss promotes tighter intra-class clustering within each modality, allowing the model to better utilize the expressive capacity of weaker modalities. Unlike approaches that rely on dominant modalities to shape representations, our method encourages each modality to refine its features based on its own semantic structure, thus alleviating modality imbalance.

\subsection{Cross-modal Alignment Learning via Entropic Optimal Transport (CAL)}
Intra-modal optimization alone is insufficient to guarantee robust cross-modal fusion, particularly due to the inherent distributional heterogeneity across modalities \cite{yang2023confede}.
To address this, we align different modalities using an Entropic Optimal Transport (EOT) framework. 

Formally, let $\{c_k^{(m)}\}_{k=1}^K$ and $\{c_l^{(n)}\}_{l=1}^K$ denote the sets of class prototypes from modalities $m$ and $n$, respectively. These prototypes are treated as discrete distributions $P^{(m)}$ and $P^{(n)}$ in the latent space. The alignment objective is to minimize the transportation cost between the two distributions, formulated as the Wasserstein distance:

\begin{equation}
\mathcal{L}_{\mathrm{proto}}^{m,n}=\inf_{\pi\in\Pi(P^m,P^n)}\mathbb{E}_{(x,y)\sim\pi}[\|x-y\|_2^2]
\label{eq:OT}
\end{equation}
where $\Pi(P^m,P^n)$ denotes the set of all joint distributions (transport plans) with marginals $P^m$ and $P^n$ , and $\pi$ defines the matching strategy between prototype pairs across modalities.
However, directly optimizing Eq.~\ref{eq:OT} is often computationally prohibitive \cite{gushchin2023entropic,kassraie2024progressive}. To this end, we adopt an entropy-regularized relaxation, known as Entropic Optimal Transport (EOT), which enables stable and tractable alignment. Given a ground cost matrix $\mathbf{C}\in\mathbb{R}^{K\times K}$ with entries $C_{kl}=d(c_k^{(m)},c_l^{(n)})$, the EOT objective is defined as:

{\small
\begin{equation}
    \min_{\mathbf{Q}\in\mathcal{U}(\mathbf{u},\mathbf{v})}\langle\mathbf{Q},\mathbf{C}\rangle_F+\lambda \mathcal H(\mathbf{Q}),\quad\mathrm{s.t.~}\mathbf{Q}\mathbf{1}_K=\mathbf{u},\mathrm{~}\mathbf{Q}^\top\mathbf{1}_K=\mathbf{v}
\end{equation}
}where $\mathbf{Q}$ is the transport matrix, $\langle\cdot,\cdot\rangle_F$ is the Frobenius inner product, and $\mathcal{H}(\mathbf{Q})=-\sum_{k,l}\mathbf{Q}_{kl}\log \mathbf{Q}_{kl}$ is the entropy regularizer. 
The marginal constraints $\mathbf{u}$ and $\mathbf{v}$ are typically set to uniform distributions, and $\mathcal{U}(\mathbf{u}, \mathbf{v})$ denotes the space of doubly stochastic matrices.
To encourage bidirectional semantic consistency, we compute the transport plans in both directions ( $\mathbf{Q}^{(m \to n)}$, $\mathbf{Q}^{(n \to m)}$ ) and define a symmetric matching loss:
{\small
\begin{equation}
    \mathcal{L}_{\mathrm{match}}=\frac{1}{2}\left(\langle \mathbf{Q}^{(m\to n)},\mathbf{C}\rangle_F+\langle \mathbf{Q}^{(n\to m)},\mathbf{C}^\top\rangle_F\right)
\end{equation}
}which promotes mutual alignment between modality pairs and alleviates one-sided matching bias.
Furthermore, to reinforce the consistency between forward and backward alignment, we introduce a consistency regularization that penalizes the discrepancy between the two transport matrices:
\begin{equation}
    \mathcal{L}_{\mathrm{reg}}=\|\mathbf{Q}^{(m\to n)}-(\mathbf{Q}^{(n\to m)})^\top\|_F^2
\end{equation}

Lastly, we incorporate a structure-preserving regularizer to encourage identity-like transport matrices, thereby preserving intra-class semantic integrity:
\begin{equation}
    \Omega(\mathbf{Q})=\|\mathbf{Q}-\mathbf{I}_K\|_F^2
\end{equation}
where $\mathbf{I}_K$ denotes the identity matrix. 
By integrating the above components, the total loss function for cross-modal prototype alignment is formulated as:
{\small
\begin{equation}
\large\mathcal{L}_{\mathrm{inter}}=\sum_{(m,n)\in\mathcal{P}}\left(\mathcal{L}_\mathrm{match}^{(m,n)}+\alpha\mathcal{L}_{\mathrm{reg}}^{(m,n)}+\beta\Omega(\mathbf{Q})\right)
\label{eq:inter_alignment}
\end{equation}
}where the hyperparameters $\alpha$ and $\beta$ control the relative importance of the consistency constraint and the structure-preserving regularization.

\subsection{Dual-Phase Optimization Module}

\subsubsection{Modality Initialization Learning Phase}
Early-stage multimodal learning often suffers from modality dominance, where stronger modalities overshadow weaker ones \cite{yu2019adapting, yang2020cm, mai2022hybrid}. To mitigate this issue, we introduce an entropy-guided modality weighting strategy, which assigns importance to each modality based on the uncertainty of its predictions:
\begin{equation}
    E(H^m) = 1 - \frac{1}{N} \sum_{i=1}^{N} p(y_i|h_i^m) \log p(y_i|h_i^m)
\end{equation}
The resulting modality weight is computed via a softmax over all modalities:
\begin{equation}
    \alpha^{(m)} = \frac{\exp(E(H^m))}{\sum_{k \in \mathcal{M}} \exp(E(H^k))}
\end{equation}

\textbf{Modality Fusion}
We employ a Prototype-Gated Fusion (PGF) module to dynamically integrate modality-specific representations. Each modality’s representation $h^m$ is modulated via a gating mechanism conditioned on both its learned embedding and corresponding class prototype and obtain the fused representation:
\begin{equation}
    g^{(m)} = \sigma(W_g^m[\alpha^{(m)} h^m \oplus c_y^m])
\end{equation}
\begin{equation}
    h_{\mathrm{fusion}} = \mathrm{FFN}\left(\bigoplus_{m \in \mathcal{M}} g^{(m)} \cdot h^m \right)
\end{equation}
The final sentiment prediction $\hat{y}$ is computed by:
\begin{equation}
    \hat{y} = \mathrm{Softmax}(\mathrm{MLP}(h_{\mathrm{fusion}}))
\end{equation}

\subsubsection{Objective Function}
The training objective integrates both the classification loss and prototype alignment regularization:
\begin{equation}
    \mathcal{L} = \mathcal{L}_\mathrm{task} + \mu \mathcal{L}_\mathrm{align}
\end{equation}
\begin{equation}
    \mathcal{L}_\mathrm{task} = \frac{1}{N} \sum_{i=1}^{N} - y_i \log(\hat{y}_i)
\end{equation}
\begin{equation}
    \mathcal{L}_\mathrm{align} = \sum_{m \in \mathcal{M}} \alpha^{(m)} \mathcal{L}_{\mathrm{intra}}^{m} + \mathcal{L}_{\mathrm{inter}}
\end{equation}
This function allows the dominant modality to guide training while improving weaker ones through alignment feedback.

\subsubsection{Modality Balance Optimization Phase}

To address the long-term imbalance in gradient updates caused by modality dominance, we propose a Shapley-based Gradient Modulation (SGM) mechanism. Inspired by cooperative game theory \cite{gul1989bargaining}, we quantify the marginal contribution of each modality using the Shapley value:
\begin{equation}
    \psi_m(u) = \sum_{S \subseteq \mathcal{M} \setminus \{m\}} \frac{|S|!(k - |S| - 1)!}{k!} \left[u(S \cup \{m\}) - u(S)\right]
\end{equation}
The utility function $u(S)$ is defined as:
\begin{equation}
    u(S) = \frac{\rho}{1 + \mathcal{L}_{\mathrm{inter}}^S} + \frac{1 - \rho}{1 + \mathcal{L}_{\mathrm{intra}}^S}
\end{equation}
where $\rho \in [0,1]$ controls the balance between inter- and intra-modal performance based on Eq.~\ref{eq:3} and Eq. ~\ref{eq:inter_alignment}. We normalize Shapley values and use them to modulate gradient updates:
\begin{equation}
    \tilde{\psi}_m = \frac{\psi_m}{\sum_k \psi_k}, \quad
    \varphi_m = \exp\left( \frac{\tilde{\psi}_{\min}}{\tilde{\psi}_m} - 1 \right)
\end{equation}
where $\varphi_m$ serves as a modulation factor that dynamically adjusts the learning rate of each modality by amplifying the updates of weaker modalities and suppressing the perturbations of dominant ones.
The parameter update of modality $m$ at iteration $t+1$ is:
\begin{equation}
    \theta_{t+1}^m = \theta_t^m - \eta \cdot \varphi_m \cdot \frac{\partial \mathcal{L}^{(t)}}{\partial \theta^m}
\end{equation}
This stage encourages optimization focus to shift toward under-performing modalities, thereby promoting a balanced fusion process.

\section{Experiments}
\subsection{Datasets and Evaluation Metrics}
To validate the effectiveness of PaSE, we conduct experiments on three benchmark datasets: CMU-MOSI \cite{zadeh2016mosi}, CMU-MOSEI \cite{bagher-zadeh-etal-2018-multimodal}, and IEMOCAP \cite{busso2008iemocap}. 

We report the averaged results over five runs with different random seeds. For classification, we use 2-class (Acc-2) and 7-class (Acc-7) accuracy for MOSI and MOSEI, and F1 for IEMOCAP. For regression, we report Mean Absolute Error (MAE) and correlation (Corr), where all metrics except MAE favor higher values. Note that Acc-2 and F1 on CMU-MOSI and CMU-MOSEI are reported in two forms (denoted as `-/-'): one includes zero as negative/non-negative, the other excludes it as negative/positive.
 
\begin{table*}[t]
\centering
\begin{adjustbox}{width=2.10\columnwidth}
\begin{tabular}{l|cccccccccccc}
\toprule
\multirow{2}{*}{Method} & \multicolumn{5}{c}{CMU-MOSI} & \multicolumn{5}{c}{CMU-MOSEI} \\
\cmidrule(lr){2-6} \cmidrule(lr){7-11}
 &  Acc-2($\uparrow$) & F1($\uparrow$) & Acc-7($\uparrow$) & Corr($\uparrow$) & MAE($\downarrow$) & Acc-2($\uparrow$) & F1($\uparrow$) & Acc-7($\uparrow$) & Corr($\uparrow$) & MAE($\downarrow$)\\
\midrule

MuLT$_{\textrm{\fontsize{6}{6}\selectfont ACL'19}}^\heartsuit$ & 79.51/80.47 & 79.46/80.49 & 36.74 & 0.667 & 0.892 &	81.10/83.63 & 81.05/83.46 & 52.34 & 0.605 & 0.671 \\
MAG-BERT$_{\textrm{\fontsize{6}{6}\selectfont ACL'20}}^\diamondsuit$ & 82.37/84.43 & 82.50/84.61 & 43.62 & 0.727 & 	0.781 & 82.51/84.82 & 82.77/84.71 & 52.67 & 0.755 & 0.543 \\
SelfMM$_{\textrm{\fontsize{6}{6}\selectfont AAAI'21}}^\diamondsuit$& 82.54/84.77 & 84.42/85.95 & 45.79 & 0.795 & 0.712 & 82.68/84.96 & 82.95/84.93 & 53.46 & 0.767 & 0.529 \\
HyCon$_{\textrm{\fontsize{6}{6}\selectfont TAC'22}}^\diamondsuit$ & -/85.20 & 82.68/84.91 & 46.60 & 0.790 & 0.713 & -/85.40 & -/85.60 & 52.80 & 0.776 & 0.601 \\
ConKI$_{\textrm{\fontsize{6}{6}\selectfont ACL'23}}^\diamondsuit$ & 84.37/86.13 & 84.33/86.13 & 48.43 & 0.816 & \underline{0.681} & 82.73/86.25 & 83.08/86.15 & 54.25 & 0.782 & 0.529 \\
ConFEDE$_{\textrm{\fontsize{6}{6}\selectfont ACL'23}}$ & 84.17/85.52 & 84.13/85.52 & 42.27 & 0.742 & 0.784 & 81.65/85.82 & 82.17/85.83 & 54.86 & 0.780 & 0.522 \\
CLGSI$_{\textrm{\fontsize{6}{6}\selectfont NAACL'24}}$& 83.97/86.43 & 83.63/86.25 & 47.96 & 0.790 & 0.703 & 84.01/86.32 & 84.21/86.18 & 54.56 & 0.763 & 0.532 \\
MCL-MCF$_{\textrm{\fontsize{6}{6}\selectfont TAC'24}}$ & 84.90/87.30 & 84.70/87.20 & - & 0.799 & 0.692 & 84.20/86.40 & 84.40/86.30 & - & 0.767 & 0.536 \\
MFON$_{\textrm{\fontsize{6}{6}\selectfont COLING'25}}$ & 84.84/86.89 & 84.75/86.86 & 44.90 & 0.797 & 0.725 & 82.70/86.32 & 83.13/86.29 & 53.72 & 0.780 & 0.528 \\
GLoMo$^\heartsuit_{\textrm{\fontsize{6}{6}\selectfont MM'25}}$ & 84.10/86.70 & 83.90/86.60 & 48.30 & 0.782 & 0.718 & 83.70/86.50 & 84.00/86.40 & 55.00 & 0.771 & 0.539 \\

EUAR$^*_{\textrm{\fontsize{6}{6}\selectfont MM'25}}$ & 83.24/84.58 & 84.22/84.58 & 48.18 & 0.789 & 0.711 & 82.20/85.64 & 82.02/85.46 & 51.07 & 0.785 & 0.596 \\

KEBR$_{\textrm{\fontsize{6}{6}\selectfont MM'25}}$ & 84.84/87.27 & 84.83/87.25 & 47.81 & 0.819 & 0.683 & 84.01/86.74 & 84.25/86.68 & 54.37 & {0.799} & 0.517 \\


Semi-IIN$_{\textrm{\fontsize{6}{6}\selectfont AAAI'25}}$ &85.28/87.04 & 85.19/87.00 & 46.50 & \underline{0.822} & \textbf{0.679} & 84.98/\underline{87.70} & 85.27/\underline{87.65} & \textbf{55.89}  & \underline{0.804} & \textbf{0.497} \\

MSAmba$_{\textrm{\fontsize{6}{6}\selectfont AAAI'25}}$ & \underline{85.99/87.43} & \underline{85.99/87.40} & \underline{49.67} & 0.809 & 0.707 & \underline{85.78}/86.86 & \underline{85.99}/86.93 & 54.21 & 0.796 & \underline{0.507} \\

\textbf{PaSE(ours)} & \textbf{86.40/88.32} & \textbf{86.34/88.25} & \textbf{50.92} & \textbf{0.847} & 0.695 & \textbf{86.07/88.10} & \textbf{86.21/87.96} & \underline{55.76} & \textbf{0.831} & 0.523 \\

\bottomrule
\end{tabular}
\end{adjustbox}
\caption{Comparison with SOTA methods on MOSI and MOSEI. Bold indicates the best result, and underlining denotes the second-best. 
Note: $\heartsuit$ and $\diamondsuit$ represents results obtained from \citet{zhuang2024glomo} and \citet{yu2023conki}, respectively.  ``-'' denotes the result is not provided, * indicates reproduced results from the public code. Other results taken directly from the original paper.}
\label{tab:main_results_mosi_mosei}

\end{table*}

\subsection{Baselines}
In this study, we select a diverse set of current SOTA baselines to conduct a comprehensive comparison.
These methods include MuLT \cite{tsai2019multimodal}, MAG-BERT \cite{rahman2020integrating}, 
SelfMM\cite{yu2021learning}, HyCon \cite{mai2022hybrid}, ConKI \cite{yu2023conki}, ConFEDE \cite{yang2023confede}, CLGSI~\cite{yang2024clgsi}, MCL-MCF~\cite{fan2024multi}, 
MFON \cite{zhang2025modal}, GLoMo \cite{zhuang2024glomo}, EUAR \cite{gao2024enhanced}, KEBR \cite{zhu2024kebr}, Semi-IIN~\cite{lin2025semi}, and MSAmba~\cite{he2025msamba}. 
See Appendix A for details.

\begin{table}[t]

\centering
\setlength{\tabcolsep}{3mm}
\begin{adjustbox}{width=1.00\columnwidth}
\begin{tabular}{l|cccc}
\toprule
\textbf{Method} & \textbf{Happy} & \textbf{Sad} & \textbf{Angry} & \textbf{Neutral} \\
\midrule
MCTN$_\textrm{~\cite{pham2019found}}$          & 83.1 & 82.8 & 84.6 & 67.7 \\
Self-MM$_\textrm{~\cite{yu2021learning}}$  & \underline{90.8} & 86.7 & \underline{88.4} & \underline{72.7} \\
SMIL$_\textrm{~\cite{ma2021smil}}$          & 86.8 & 85.2 & 84.9 & 68.9 \\

GCNet$_\textrm{~\cite{lian2023gcnet}}$         & 87.7 & 86.9 & 85.2 & 71.1 \\
UMDF$_\textrm{~\cite{li2024unified}}$          & 87.9 & 86.5 & 85.8 & 70.5 \\
DMD$_\textrm{~\cite{li2023decoupled}}$         & 91.1 & 88.4 & 88.6 & 72.2 \\
CorrKD$_\textrm{~\cite{li2024correlation}}$    & 87.5 & 85.9 & 86.1 & 71.5 \\
\textbf{PaSE (ours)}               & \textbf{91.5} & \textbf{88.6} & \textbf{89.4} & \textbf{73.2} \\
\bottomrule
\end{tabular}
\end{adjustbox}
\caption{Comparison on IEMOCAP dataset across different emotion classes (F1-score, $\uparrow$).}
\label{tab:iemocap_emotions}

\end{table}

\subsection{Implementation Details}
To ensure fair comparison with existing SOTA methods, we adopt the standard feature extractors: Facet for visual, COVAREP \cite{degottex2014covarep} for acoustic, and BERT$_\text{base}$ \cite{devlin2019bert} for textual modalities on CMU-MOSI and CMU-MOSEI.
We use Adam optimizer with a learning rate of 1e-5, a batch size of 64, and train for 200 epochs. We set $\gamma=0.98$ for prototype updates, with $\lambda=0.01$, $\mu=0.1$, and alignment-enhancing factors $\alpha=0.1$, $\beta=0.05$ to improve cross-modal consistency. All experiments are conducted on a NVIDIA A100 GPU. 
A warm-up phase is first employed to stabilize modality-specific learning and prevent early overfitting. Training then automatically transitions to the Shapley-guided balancing phase once the validation entropy stabilizes, with validation accuracy monitored to ensure performance remains stable.
More details are provided in Appendix B.

\section{Results and Analysis}
\subsection{Comparison with SOTA methods}

As shown in Table~\ref{tab:main_results_mosi_mosei} and Table~\ref{tab:iemocap_emotions}, we conduct extensive evaluations on three widely-used benchmarks: CMU-MOSI, CMU-MOSEI, and IEMOCAP. Our proposed PaSE, consistently outperforms existing SOTA approaches across all datasets and most evaluation metrics.
Notably, on CMU-MOSI, PaSE surpasses the most recent SOTA model, MSAmba, achieving relative gains of 0.89\% in Acc-2 and 1.25\% in Acc-7. Comparable consistent gains are also observed on CMU-MOSEI, highlighting the effectiveness of the prototype-guided alignment mechanism in capturing modality-consistent semantics and mitigating modality noise.

On the IEMOCAP dataset, PaSE achieves the highest performance across all four emotion categories, with an average improvement of 2.93\% over prior methods. This substantial performance gain reflects the superior discriminative capability and resilience of our model in more complex emotion recognition settings. 
We attribute these improvements to the explicit alignment of modality-specific features toward shared semantic prototypes, which enhances intra-class compactness and inter-class separability in the fused representation space.

\begin{table}[t!]
\centering

\begin{adjustbox}{width=1.0\columnwidth}
\begin{tabular}{l|cc|cc|c}
\toprule
\multirow{2}{*}{Model} & \multicolumn{2}{c}{\textbf{CMU-MOSI}} & \multicolumn{2}{c}{\textbf{CMU-MOSEI}} & \multicolumn{1}{c}{\textbf{IEMOCAP}}  \\
\cmidrule(lr){2-3} 
\cmidrule(lr){4-5}  
\cmidrule(lr){6-6} 
 & Acc-2($\uparrow$) & F1($\uparrow$)  & Acc-2($\uparrow$) & F1($\uparrow$) & WA($\uparrow$) \\
\midrule
    \multicolumn{6}{c}{\textit{Ablation Study}} \\
    \midrule
    \textbf{PaSE} & \textbf{86.40/88.32} & \textbf{86.34/88.25} & \textbf{86.07/88.10} & \textbf{86.21/87.96} & \textbf{80.47} \\
    - PCL & 85.21/87.25 & 85.17/87.16 & 84.73/86.91 & 84.96/86.83 & 78.54 \\
    - CAL & 85.04/86.80 & 85.06/86.82 & 84.54/86.69 & 84.47/86.62 & 78.13 \\
    - SGM & 83.65/85.47 & 83.60/85.39 & 83.13/85.03 & 83.46/85.08 & 76.80 \\
    \midrule
    
    \multicolumn{6}{c}{\textit{Ablation on Fusion Strategy}} \\
    \midrule
    - PGF & 84.57/86.37 & 82.48/86.32 & 83.06/86.05 & 84.45/85.90 & 77.94 \\
    \ \ + Sum & 84.45/86.92 & 84.34/86.26 & 83.73/86.59 & 83.89/86.53 & 78.43 \\
    \ \ + Con & 85.92/87.40 & 85.85/87.37 & 85.37/87.14 & 85.51/87.10 & 79.16 \\
    \ \ + Att & 85.60/87.26 & 85.53/87.19 & 85.25/86.85 & 85.40/86.97 & 78.75 \\

    \bottomrule
    
\end{tabular}
\end{adjustbox}
\caption{Ablation studies for each module in our proposed model PaSE on MOSI, MOSEI and IEMOCAP datasets. We adopt weighted accuracy (WA) as evaluation metrics for IEMOCAP datasets.}
\label{tab:ablation_fusion}
\end{table}

\subsection{Ablation Study}
To evaluate the contribution of each component in PaSE, we conduct a series of ablation experiments on the MOSI and MOSEI datasets. As shown in Table ~\ref{tab:ablation_fusion}, removing the prototype-guided calibration (PCL) leads to a clear drop in performance, with Acc-2 decreasing by 1.07\% on MOSI and 1.19\% on MOSEI. This highlights the importance of the prototype-based structure in enforcing intra-class compactness and inter-class separability within modalities, which are essential for robust sentiment prediction. 
Further, removing the cross-modal alignment (CAL) module causes Acc-2 to decline by 1.52\% and 1.41\% on the two datasets, respectively, demonstrating its role in facilitating semantic coherence prior to fusion. 
Lastly, the Shapley-based gradient modulation (SGM) contributes most significantly; its removal results in 2.85\% and 3.07\% performance drop on MOSI and MOSEI, underscoring its ability to mitigate modality dominance and enhance collaborative optimization in the later training stages. 
These results collectively affirm the effectiveness and necessity of each component in PaSE.

\subsection{Ablation on Fusion Strategy}
To evaluate the effectiveness of our proposed Prototype-Gated Fusion (PGF) mechanism, we conduct a comparative analysis with several alternative fusion strategies, including `Sum' (simple addition), `Con' (MLP-based concatenation), and `Att' (attention-based fusion). 
As shown in Table~\ref{tab:ablation_fusion}, the `Sum' fails to capture modality-specific semantics, leading to significant information loss and degraded performance.
Although `Con' and `Att' offer stronger representational capacity, they struggle to resolve modality competition effectively.
In contrast, PGF explicitly accounts for the varying contributions of different modalities and enables more discriminative and balanced feature fusion, consistently outperforming baseline methods across multiple evaluation metrics.

\begin{figure}[t]
    \centering \includegraphics[width=1.00\columnwidth]{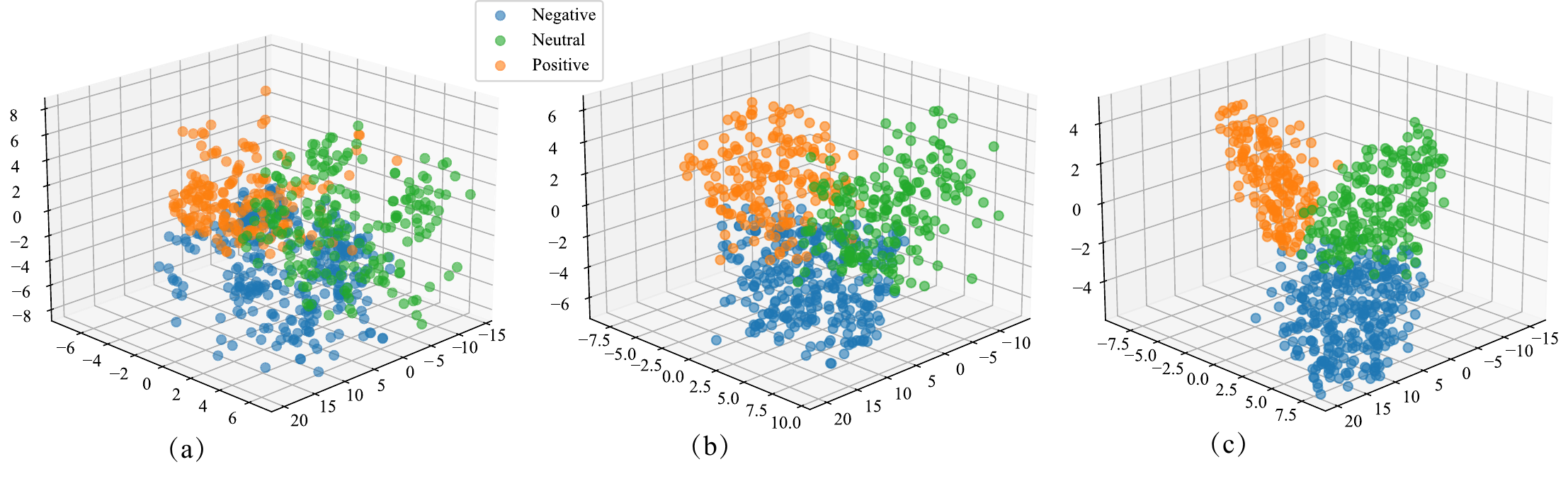}
    \caption{
    t-SNE visualization of fused feature representations obtained using different models on MOSI test set. (a) SelfMM, (b)EUAR, and (c) PaSE (ours).}
    \label{fig:tsne}
\end{figure}


\begin{table}[t]
\centering

\begin{adjustbox}{width=1.0\columnwidth}
\begin{tabular}{c|ccccc}
\toprule
\multirow{2}{*}{Model} & \multicolumn{2}{c}{\textbf{CMU-MOSI}} & \multicolumn{2}{c}{\textbf{CMU-MOSEI}} & \multicolumn{1}{c}{\textbf{IEMOCAP}}  \\
\cmidrule(lr){2-3} \cmidrule(lr){4-5}  \cmidrule(lr){6-6}
 & Acc-2($\uparrow$) & F1($\uparrow$)  & Acc-2($\uparrow$) & F1($\uparrow$) & WA($\uparrow$)\\
 
    \midrule
    A & 60.01 & 58.79 & 69.04 & 68.56 & 67.34 \\
    V & 61.43 & 61.50 & 68.69 & 68.74 & 57.31 \\
    T & 84.70 & 84.23 & 84.36 & 84.08 & 74.80 \\
    A+V & 63.35 & 63.32 & 72.03 & 71.90 & 68.75 \\
    T+A & 86.71 & 86.79 & 86.47 & 86.42 & 78.93 \\
    T+V & 87.14 & 86.99 & 86.73 & 86.45 & 77.52  \\
    \textbf{PaSE(T+A+V)} & \textbf{88.32} & \textbf{88.25} & \textbf{88.10} & \textbf{87.96} & \textbf{80.47} \\
    \midrule
    $\Delta(\%)$ & $\uparrow$\textit{3.62} & $\uparrow$\textit{4.02} & $\uparrow$\textit{3.74} & $\uparrow$\textit{3.88} & $\uparrow$\textit{5.67}  \\
    \bottomrule
    
\end{tabular}
\end{adjustbox}
\caption{Performance comparison under different modality combinations across three benchmarks. Only `negative/positive' results are reported for Acc-2 and F1, where T, A, and V denote Text, Audio, and Video, respectively. }
\label{tab:modalities_gain}
\end{table}


\subsection{Modality Fusion Gain Evaluation}
To systematically assess PaSE’s ability to alleviate modality competition and enhance multimodal synergy, we evaluate its performance under different modality combinations across three datasets.
As shown in Table~\ref{tab:modalities_gain}, incorporating additional modalities consistently improves performance, indicating PaSE’s capacity to exploit cross-modal complementarity.
Specifically, as illustrated in Figure~\ref{fig:intro_compare}, fusing the visual modality (V+T’) yields a 2.76\% improvement—significantly surpassing GCNet’s 0.60\% under the same condition. Furthermore, under the full-modality setting (T+A+V’), PaSE achieves an average gain of 4.02\% over all bimodal variants, effectively avoiding the diminishing returns observed in prior approaches such as MCTN~\cite{pham2019found} and MMIN~\cite{fernando2021missing}.
Notably, the results in Table~\ref{tab:modalities_gain} also demonstrate PaSE’s \textbf{robustness under missing-modality scenarios}, where the model maintains competitive performance even with partial inputs, confirming its strong generalization and resilience to modality absence.

\begin{figure}[t]
    \centering \includegraphics[width=0.95\columnwidth]{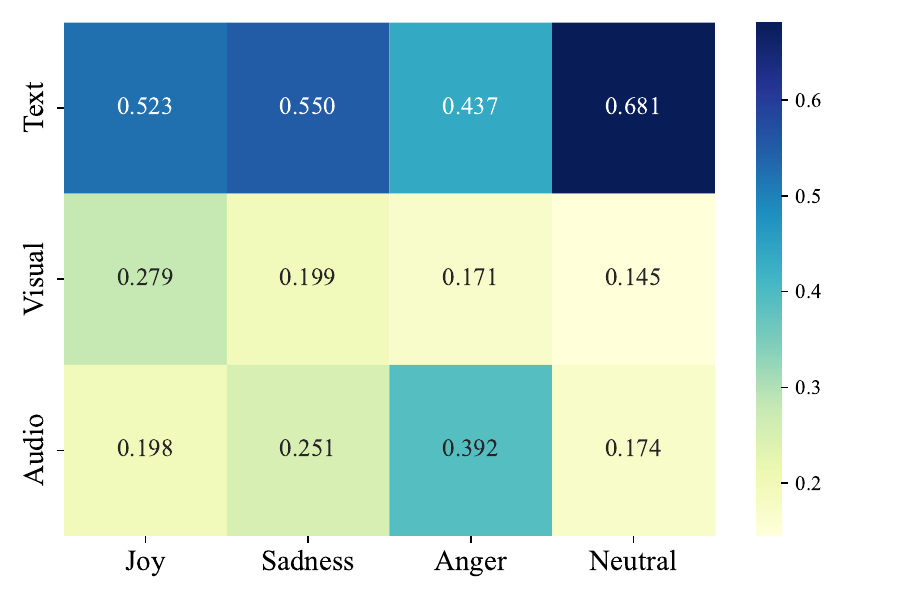}
    \caption{
    Heatmap of modality contributions for four emotion categories on IEMOCAP. Text, Visual, and Audio modalities show varying influence, with Text generally dominating, especially in Neutral and Sadness.}
    \label{fig:pgf_heatmap}

\end{figure}


\subsection{In-depth Analysis}
To further investigate the effectiveness of PaSE, we conduct in-depth analyses by exploring the following key questions:

\paragraph{\textbf{Q1. How does prototype-guided alignment enhance the discriminative structure of the fused feature space?}}
We visualize the multimodal representations learned by PaSE on the CMU-MOSI test set using t-SNE, and compare them with SelfMM \cite{yu2021learning} and EUAR \cite{gao2024enhanced}. Sentiment labels are categorized as \textit{Positive}, \textit{Neutral}, and \textit{Negative}.
As shown in Figure~\ref{fig:tsne}, the fused representations of SelfMM and EUAR exhibit substantial overlap across sentiment categories, leading to indistinct boundaries and weak intra-class cohesion. In contrast, PaSE forms compact and clearly separated clusters, indicating stronger discriminability in the fused space.
This improvement stems from PaSE’s dual alignment strategy: prototype calibration enhances intra-modality discrimination by refining class-specific structures, while entropic optimal transport enforces cross-modal semantic consistency. Jointly, these mechanisms organize representations around well-defined semantic prototypes, resulting in improved class separability and robust sentiment prediction.

\paragraph{\textbf{Q2. How does the Prototype-Gated Fusion module influence modality fusion?}}
We perform a fine-grained analysis of modality attention weights across different emotion categories on the IEMOCAP dataset. Specifically, we group the test samples by emotion class and compute the average attention weight assigned to each modality by the Prototype-Gated Fusion (PGF) module.
As illustrated in Figure~\ref{fig:pgf_heatmap}, the PGF module demonstrates strong context-aware fusion behavior by dynamically adjusting modality contributions based on the emotional semantics of each instance. 

For example, in the case of Anger, the model assigns the highest average weight to the audio modality (0.392), effectively capturing prosodic and paralinguistic cues such as pitch, intensity, and vocal tension—hallmarks of emotionally intense speech. 
In contrast, for Neutral expressions, the text modality receives the dominant weight (0.681), indicating that verbal semantics play a more critical role when emotional cues are less pronounced.
This context-sensitive behavior aligns well with human perceptual intuition, highlighting PGF’s ability to prioritize the most informative modality for each emotional state.

\begin{figure}[t]
    \centering \includegraphics[width=0.99\columnwidth]{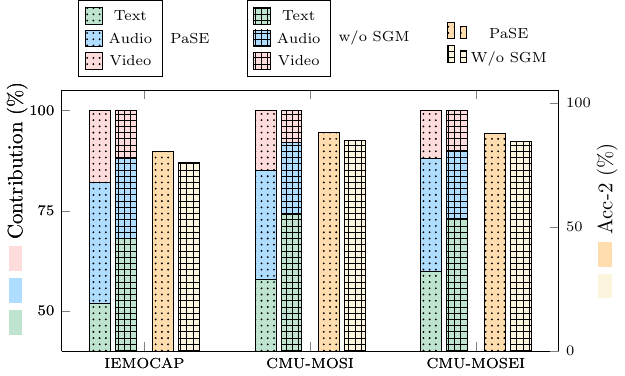}
    \caption{
   The influence of SGM on different modalities and the overall performance. The dual y-axis shows the average contribution of each modality (left) and the corresponding Acc-2 score (right).}
    \label{fig:contribution}
\end{figure}

\paragraph{\textbf{Q3. How does Shapley-guided gradient modulation alleviate modality competition?}}
We analyzed the effect of the SGM module on three benchmark datasets. 
Figure~\ref{fig:contribution} illustrates the comparison of modality contributions and the overall Acc-2 score. 
Without SGM, the text modality overwhelmingly dominates the fusion process, indicating strong modality competition and potential overfitting to a single modality.
After incorporating SGM, the dominance of the text modality is significantly reduced. Contributions from the visual and audio modalities increase correspondingly, resulting in a more balanced and cooperative fusion process. 
Furthermore, the overall fusion accuracy improves consistently during training when SGM is applied. These results validate that SGM effectively alleviates modality competition by encouraging cooperative gradient interactions among modalities. Consequently, this leads to improved modality balance and enhances the robustness and generalization ability of the multimodal fusion model.

\paragraph{\textbf{Q4. Do general-purpose LLMs generalize well enough for multimodal sentiment analysis tasks, or is specialized adaptation still necessary?}}

We conduct comparative experiments to evaluate the effectiveness of LLMs and multimodal LLMs (MLLMs) on multimodal sentiment analysis.
While these models demonstrate impressive generalization across a wide range of NLP tasks, they \textit{underperform in domain-specific scenarios}, where modality coordination and emotional nuance play a critical role.
As shown in Table~\ref{tab:results_with_LLMs}, LLM-based approaches achieve substantially lower performance than specialized multimodal architectures, particularly on the IEMOCAP dataset, which involves spontaneous dialogue, speaker variation, and subtle emotional ambiguity.
These results highlight a fundamental limitation of general-purpose models: without dedicated adaptation or modality-aware optimization, they fail to capture the fine-grained inter-modal dependencies essential for robust affective reasoning.
In contrast, PaSE, built upon a lightweight BERT-base backbone, consistently outperforms these general-purpose models across all benchmarks.
By incorporating prototype-guided alignment and Shapley-based optimization, PaSE introduces task-specific inductive biases that strengthen cross-modal interaction, improve semantic consistency, and yield more stable and interpretable affective representations.

\begin{table}[t!]
\centering

\begin{adjustbox}{width=1.0\columnwidth}
\begin{tabular}{l|lccc}
\toprule
& \multirow{2}{*}{\textbf{Models}} & \multicolumn{1}{c}{\textbf{IEMOCAP}} & \multicolumn{1}{c}{\textbf{MOSI}} & \multicolumn{1}{c}{\textbf{MOSEI}}  \\

&  & \textbf{WA(\%)} & \textbf{Acc-2(\%)} & \textbf{Acc-2(\%)} \\
\midrule
\multirow{3}{*}{\textbf{LLMs}} 
& Qwen2-7B         & 40.15 & 81.69 & 80.80 \\
& Llama-3-8B       & 38.60 & 81.44 & 80.71 \\
& Internlm-2.5-7B  & 39.49 & 84.27 & 82.95 \\
\midrule
\multirow{5}{*}{\textbf{MLLMs}} 
& mPLUG-Owl        & \underline{55.17} & 72.33 & 72.21 \\
& Qwen2-VL-7B      & 34.23 & 79.84 & 79.39 \\
& LLaVA-OV-7B      & 40.09 & \underline{87.10} & \underline{86.73} \\
& VideoLLaMA2-7B   & 38.86 & 79.17 & 78.54 \\
& GPT-4o-mini      & 46.33 & 86.54 & 86.06 \\
\midrule
\textbf{Bert-base} & \textbf{PaSE (Ours)} & \textbf{80.47} & \textbf{88.32} & \textbf{88.10} \\
\bottomrule
\end{tabular}
\end{adjustbox}
\caption{Comparison of our PaSE model with several LLMs and MLLMs on IEMOCAP, MOSI, and MOSEI datasets. Metrics are Weighted Accuracy (WA) for IEMOCAP and binary accuracy (Acc-2) for MOSI and MOSEI.}
\label{tab:results_with_LLMs}
\end{table}

\section{Conclusion}

This paper proposes the PaSE framework to dynamically balance modality collaboration and competition in multimodal sentiment analysis. 
We utilize prototype calibration to perform intra-modality alignment, strengthening the representation of non-dominant modalities while preventing their suppression. 
For cross-modality alignment, an entropic optimal transport mechanism is employed to achieve effective semantic alignment between modalities. 
Furthermore, a dual-phase optimization strategy is designed: the first phase adaptively fuses features based on modality contributions to enhance overall representation, and the second phase applies Shapley-based gradient modulation to mitigate dominance-induced suppression. 
Extensive experiments on CMU-MOSI, CMU-MOSEI and IEMOCAP datasets, together with thorough ablation studies, validate the robustness and efficacy of PaSE and its core components.

\section{Acknowledgments}
This work is supported by the National Natural Science Foundation of China (No. 62176187).

\bibliography{aaai2026}

\clearpage
\appendix

\renewcommand\thefigure{A\arabic{figure}}
\setcounter{figure}{0}
\renewcommand\thetable{A\arabic{table}}
\setcounter{table}{0}
\renewcommand\theequation{A\arabic{equation}}
\setcounter{equation}{0}
\pagenumbering{arabic}
\renewcommand*{\thepage}{A\arabic{page}}

\makeatletter
\setcounter{secnumdepth}{2} 
\renewcommand{\thesection}{\Alph{section}} 
\renewcommand{\thesubsection}{\Alph{section}.\arabic{subsection}} 
\makeatother

\twocolumn[
\vspace*{1em}
\begin{center}
{\LARGE\bfseries
\setlength{\baselineskip}{1.4\baselineskip}%
PaSE: Prototype-aligned Calibration and Shapley-based Equilibrium for Multimodal Sentiment Analysis
}
\end{center}
\vspace{3em}
]

\section{Baseline Models}
In this paper, we select a diverse set of current state-of-the-art baselines to conduct a comprehensive comparison, which includes:
\begin{itemize}
    \item MuLT \cite{tsai2019multimodal}: proposes a Multimodal Transformer with directional crossmodal attention to model unaligned multimodal sequences by adaptively attending to interactions across modalities and time steps without explicit alignment.
    \item MAG-BERT \cite{rahman2020integrating}: proposes the Multimodal Adaptation Gate to enhance BERT and XLNet for multimodal language tasks by dynamically adjusting their internal representations based on visual and acoustic inputs during fine-tuning, achieving human-level sentiment analysis performance.
    \item SelfMM\cite{yu2021learning}: proposes a self-supervised framework, the model generates unimodal labels to enable joint training of multimodal and unimodal tasks. A dynamic weight-adjustment strategy prioritizes samples with conflicting modality supervisions, enhancing the capture of nuanced differences.
    \item HyCon \cite{mai2022hybrid}: employs an adapter-based architecture to inject domain-specific knowledge into general pretrained models, enabling joint learning of modality-specific and universal representations
    \item ConKI \cite{yu2023conki}: employs an adapter-based architecture to inject domain-specific knowledge into general pretrained models, enabling joint learning of modality-specific and universal representations.
    \item ConFEDE \cite{yang2023confede}: introduces contrastive feature decomposition to split text, audio, and visual modalities into similarity and dissimilarity components, guided by text-centric contrastive learning.
    \item CLGSI~\cite{yang2024clgsi}: introduces a sentiment intensity-guided contrastive learning approach with weighted sample selection and a novel GLFK fusion mechanism for effective multimodal feature extraction.
    \item MCL-MCF~\cite{fan2024multi}: proposes a multi-level contrastive learning and multi-layer convolution fusion framework, which progressively mitigates modality heterogeneity through three-level contrastive learning (unimodal, cross-modal, and high-level fusion) and enhances feature fusion via tensor convolution.
    \item MFON \cite{zhang2025modal}: introduces a Modal Feature Optimization Network (MFON) with a Modal Prompt Attention (MPA) mechanism, identifying under-optimized modalities and focusing on their features via task-specific prompts.
    \item GLoMo \cite{zhuang2024glomo}: uses modality-specific mixture of experts layers to integrate diverse local representations within each modality and a global-guided fusion module to effectively combine global and local information
    \item EUAR \cite{gao2024enhanced}: introduces the Enhanced Experts with Uncertainty-Aware Routing (EUAR), integrating the Mixture of Experts approach to dynamically adapt the network via conditional computation, refining experts to capture data uncertainty and using a U-loss-based routing mechanism to direct samples to low-uncertainty experts for noise-free feature extraction in multimodal sentiment analysis.
    \item KEBR \cite{zhu2024kebr}: utilizes text-based cross-modal fusion to inject non-verbal information into text representations, applying a multimodal cosine constrained loss to balance joint learning
    \item Semi-IIN~\cite{lin2025semi}: proposes a semi-supervised intra-modal and inter-modal interaction learning network for multimodal emotion analysis. Semi-IIN integrates a masked attention mechanism and a gating mechanism, enabling effective dynamic selection after independently capturing intra-modal and inter-modal interaction information.
    \item MSAmba~\cite{he2025msamba}: composes an Intra-modal Sequential Mamba (ISM) module and a Cross-modal Hybrid Mamba (CHM) module to enhance cross-modal interactions.
\end{itemize}

\section{Implementation Details}

To prevent overfitting in the first stage of our dual-phase training, we employ a warm-up period to ensure stable modality-specific learning. When validation performance plateaus, the training automatically transitions to the second phase, activating Shapley-based modality balancing.
To ensure fairness, we adopt a feature extraction framework aligned with existing SOTA methods. 
Specifically, we use Facet, COVAREP \cite{degottex2014covarep}, and BERT \cite{devlin2019bert} as the feature extractors for the visual, acoustic, and textual modalities, respectively, on the CMU-MOSI and CMU-MOSEI datasets.
We use Adam optimizer with a learning rate of 1e-5, a batch size of 64, and train for 200 epochs. We set $\gamma=0.98$ for prototype updates, with $\lambda=0.01$, $\mu=0.1$, and alignment-enhancing factors $\alpha=0.1$, $\beta=0.05$ to improve cross-modal consistency.  
All experiments are conducted on a single NVIDIA A100 GPU.
To mitigate overfitting in the first phase of our proposed dual-phase training strategy, we introduce a warm-up period during the initial training phase, allowing sufficient learning of modality-specific representations. 
If task performance on the validation set does not significantly improve for several consecutive epochs, the model automatically transitions to the second phase, where a Shapley value-based modality balancing mechanism is activated.

\begin{table}[t]
\centering
\begin{tabular}{l|c|c|c}
\toprule
\textbf{Dataset} & \textbf{\# Train} & \textbf{\# Validation} & \textbf{\# Test} \\
\midrule
CMU-MOSI  & 1,284  & 229   & 686   \\
CMU-MOSEI & 16,326 & 1,861 & 4,659 \\
IEMOCAP & 2,717 & 798 & 938 \\

\bottomrule
\end{tabular}
\caption{Datasets statistics for fine-tuning and testing}
\label{tab:dataset_statistics}
\end{table}

\section{Details of Datasets}
PaSE is evaluated on multiple tasks, including multimodal sentiment analysis and multimodal emotion recognition, using three widely adopted benchmark datasets: CMU-MOSI, CMU-MOSEI, and IEMOCAP. 
As summarized in Table~\ref{tab:dataset_statistics}, detailed statistics and descriptions of these datasets are provided below.

\paragraph{CMU-MOSI:}consists of 2,199 single-person short video clips, with the training set (1,284 samples), validation set (229 samples), and test set (686 samples) configured according to the standard division. 

\paragraph{CMU-MOSEI:}contains 22,856 movie review videos from YouTube, divided according to the widely accepted academic standard: 16,326 training samples, 1,871 validation samples, and 4,659 test samples. 
Both datasets use manually annotated labels, with sentiment intensity scores on a continuous scale from -3 to 3, corresponding to five levels of sentiment intensity: "highly negative - negative - neutral - mildly positive - highly positive." 

\paragraph{IEMOCAP:}consists of 4,453 video clips, including 2,717 training samples, 798 validation samples, and 938 test samples. We adopt four emotion categories (\textit{Happy}, \textit{Sad}, \textit{Angry}, and \textit{Neutral}) for sentiment recognition. For evaluation, we report the F1 score for each category.

\section{More Details of Evaluation Metrics}
We report both classification and regression results, averaged over five runs with different random seeds. 
For classification tasks, we provide multiclass accuracy and F1 scores. 
Specifically, for the CMU-MOSI and CMU-MOSEI datasets, we evaluate both 2-class (Acc-2) and 7-class (Acc-7) accuracy. For IEMOCAP dataset, we provide F1-scores. 
For regression tasks, we report the Mean Absolute Error (MAE) and the Correlation (Corr). Except for MAE, higher values indicate better performance.
For the MOSI and MOSEI datasets, the Acc-2 and F1-score have two forms using the segmentation marker `-/-': 
the first represents negative/non-negative (including zero) and the second represents negative/positive (excluding zero).

\begin{figure}[t]
    \centering \includegraphics[width=0.95\columnwidth]{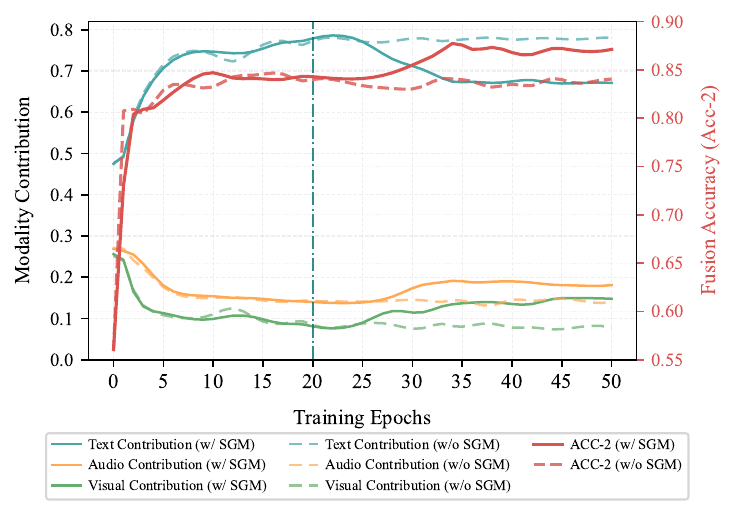}
    \caption{
   Performance evaluation of modality contribution and fusion ACC-2 on CMU-MOSI. $w/o ~SGM$: no modulation is applied throughout training, following conventional methods; $w/ ~SGM$: the SGM module is introduced starting from epoch 20. }
    \label{fig:intro_demo}
\end{figure}

\section{Effectiveness of Dual-Phase Optimization}
In most multimodal sentiment analysis (MSA) tasks, the text modality naturally holds an informational advantage, which often leads to the dominant modality suppressing the expressive capacity of other modalities during training. 
PaSE adopts a dual-phase training strategy: the first phase focuses on constructing clear semantic structures for each modality, while the second phase introduces Shapley-based gradient modulation (SGM) to adjust modality contributions. 
This design is based on a key observation: introducing Shapley-based modulation too early—before stable modality representations are formed—may lead to unstable optimization.

Figure \ref{fig:intro_demo} illustrates the effectiveness of this dual-phase mechanism. 
The left Y-axis in the figure represents the contributions of each modality, while the red right Y-axis indicates the Acc-2 performance. 
The red dashed line corresponds to the full training process without SGM, and the red solid line shows the result of introducing SGM after the 20th epoch. 
As shown, the model nearly converges within the first 20 epochs; after introducing SGM, the contribution of the text modality slightly decreases, but the overall performance improves by approximately 4\%. 
SGM addresses this issue by leveraging Shapley values to measure the marginal contribution of each modality. 
In the later stages of training, it applies moderate gradient suppression to the dominant modality (text), thereby increasing the participation of weaker modalities. 
This leads to more balanced and generalizable fused representations, ultimately improving overall performance.

\begin{table}[t]
\centering
\begin{tabular}{l|c|c}
\toprule
\textbf{Model} & \textbf{Training Time} & \textbf{Params} \\
\midrule
SelfMM & 7.7h & 121,835,723 \\
GLoMo  & 9.3h & 109,818,887 \\
EUAR   & 3.9h & 110,436,422 \\
KEBR   & 4.6h & 127,467,631 \\
PaSE   & 4.0h & 114,795,953 \\
\bottomrule
\end{tabular}
\caption{Comparison of training efficiency and parameters across different MSA models.}
\label{tab:training_efficiency}
\end{table}

\section{Efficiency Analysis}
To better evaluate the efficiency of our model, we analyze the computational complexity of its key components. PaSE consists of three modules: PCL, with complexity $O(B\times d)$; CAL, with complexity $O(B \times d \times \log d)$, attributed to the use of Entropic Optimal Transport for cross-modal alignment; and SGM, with complexity $O(M \times B)$, where $B$ is the batch size, $d$ is the feature dimension, $K$ is the number of classes, and $M$ is the number of modalities. 
Overall, the time complexity of PaSE remains linear with respect to input size (i.e., $O(n)$) and is computationally tractable. 
As shown in Table \ref{tab:training_efficiency}, it is comparable to existing MSA methods such as SelfMM, ConKI, and KEBR, demonstrating that the improved performance of PaSE does not come at the cost of excessive computational overhead.

\end{document}